\documentclass[twoside,11pt]{article}
\usepackage{jmlr2e}
\usepackage{multirow}
\usepackage[hypcap=false]{caption}

\jmlrheading{1}{2021}{1-48}{9/22; Revised 9/22}{9/22}{meila00a}{Panagiotis Anagnostou, Sotiris Tasoulis, Vassilis Plagianakos, Dimitris Tasoulis}

\ShortHeadings{HiPart: Hierarchical Divisive Clustering Toolbox}{Anagnostou, Tasoulis, Plagianakos and Tasoulis}
\firstpageno{1}

\begin{document}

\title{HiPart: Hierarchical Divisive Clustering Toolbox}

\author{\name Panagiotis Anagnostou \email panagno@uth.gr%
        \AND
        \name Sotiris Tasoulis \email stasoulis@uth.gr%
        \AND
        \name Vassilis Plagianakos \email vpp@uth.gr                    \\
        \addr Department of Computer Science and Biomedical Informatics \\
        University of Thessaly                                          \\
        Papasiopoulou 2-4, Lamia 35131, Greece
        \AND
        \name Dimitris Tasoulis \email d.tasoulis@thesignalgroup.com    \\
        \addr Signal Ocean SMPC                                         \\
        Leof. Vouliagmenis 110, Athina 11743, Greece
        }

\editor{Kevin Murphy and Bernhard Sch{\"o}lkopf}

\maketitle

\begin{abstract}%
This paper presents the HiPart package, an open-source native python library that provides efficient and interpret-able implementations of divisive hierarchical clustering algorithms. HiPart supports interactive visualizations for the manipulation of the execution steps allowing the direct intervention of the clustering outcome. This package is highly suited for Big Data applications as the focus has been given to the computational efficiency of the implemented clustering methodologies. The dependencies used are either Python build-in packages or highly maintained stable external packages. The software is provided under the MIT license. The package's source code and documentation can be found at \url{https://github.com/panagiotisanagnostou/HiPart}.
\end{abstract}

\begin{keywords}
    Clustering, High dimensionality, Python
\end{keywords}

\section{Introduction}

A highly researched problem by a variety of research communities is the problem of data clustering. However, high-dimensional data clustering still constitutes a significant challenge, plagued by the \emph{curse of dimensionality} \citep{hutzenthaler2020overcoming}. Hierarchical divisive algorithms developed in the recent years \citep{TASOULIS20103391, pavlidis2016minimum, hofmeyr2016clustering, hofmeyr2019minimum, hofmeyr2019ppci} have shown great potential for the particular case of high dimensional data, incorporating dimensionality reduction iteratively within their algorithmic procedure. Additionally, they seem unique in providing a hierarchical format of the clustering result with low computational cost, in contrast to the commonly used but computationally demanding agglomerative clustering methods.

Although the discovery of a hierarchical format is crucial in many fields, such as bioinformatics \citep{luo2003hierarchical,modena2014gene}, to the best of our knowledge, this package is the first native python implementation of divisive hierarchical clustering algorithms. We particularly focus on the ``Principal Direction Divisive Clustering (PDDP)'' algorithm \citep{boley1998principal} for its potential to effectively tackle the \emph{curse of dimensionality} and its impeccable time performance \citep{TASOULIS20103391}.

Simultaneously, we provide implementations of a complete set of hierarchical divisive clustering algorithms with a similar basis. These are the dePDDP \citep{TASOULIS20103391}, the iPDDP \citep{TASOULIS20103391}, the kM-PDDP \citep{zeimpekis2008principal}, and the bisecting k-Means (BKM) \citep{savaresi2001performance}. We also provide additional features not included in the original developments of the aforementioned methodologies that make them appropriate for the discovery of arbitrary shaped or non-linear separable clusters. In detail, we incorporate kernel Principal Component Analysis (kPCA) \citep{Scholkopf99kernelprincipal} and Independent Component Analysis (ICA) \citep{hyvarinen2000independent, tharwat2020independent} for the iterative dimensionality reduction steps. 

As a result, the package provides a fully parameterized set of algorithms that can be applied in a diverse set of applications, for example, non-linear separable clusters, automated identification for the cluster number, and outlier control.

\section{Software Description}

The HiPart (Hierarchical Partitioning) package is divided into three major sections:
\begin{itemize}
    \setlength{\itemsep}{5pt}
    \setlength{\parskip}{0pt}
    \setlength{\parsep}{0pt}
    \item Method implementation
    \item Static Visualization
    \item Interactive Visualization
\end{itemize}

\subsection{Method Implementation}

The package employs an object-oriented approach for the implementation of the algorithms, similarly to that of~\cite{JMLR:v23:21-0862}, while incorporating design similarities with the scikit-learn library \citep{pedregosa2011scikit}. Meaning, a class executes each of the algorithms, and the class's parameters and attributes are the algorithm's hyper-parameters and results.

For the execution of the algorithms, the user needs to call either the method \textbf{predict} or \textbf{fit\_predict} of each algorithm's execution class. The algorithm parameterization can be applied at the constructor of their respective class.

\subsection{Static Visualization}

Two static visualization methods are included. The first one is a 2-Dimensional representation of all the data splits generated by each algorithm during the hierarchical procedure. The goal is to provide an insight to the user regarding each node of the clustering tree and, subsequently, each step of the algorithm's execution. 

The second visualization method is a dendrogram that represents the splits of all the divisive algorithms. The dendrogram's figure creation is implemented by the \emph{SciPy} package, and it is fully parameterized as stated in the library.

\subsection{Interactive Visualization}

In the interactive mode, we provide the possibility for stepwise manipulation of the algorithms. The user can choose a particular step (node of the tree) and manipulate the split-point on top of a two-dimensional visualization, instantly altering the clustering result. Each manipulation resets the algorithm's execution from that step onwards, resulting in a restructuring of the sub-tree of the manipulated node.

\section{Development Notes}

For the development of the package, we complied with the \textbf{PEP8} style standards, and we enforced it with the employment of \emph{flake8} command-line utility. To ensure the code's quality, we implemented the \emph{unittest} module to the entirety of the source code. In addition, platform compatibility has been assured through extensive testing, and the package development in its entirety uses only well-established or native python packages. The package has been released as open-source software under the ``MIT License''. For more information regarding potential contributions or for the submission of an issue, or a request, the package is hosted as a repository on Github.

\section{Experiments and Comparisons}

In this section, we provide clustering results with respect to the execution speed and clustering performance for the provided implementations. For direct comparison, we employ a series of well-established clustering algorithms. These are the k-Means \citep{likas2003global}, the Agglomerative (AGG) \citep{ackermann2014analysis} and the OPTICS \citep{ankerst1999optics} of the scikit-learn \citep{pedregosa2011scikit} python library and the fuzzy c-means (FCM) algorithm \citep{bezdek1984fuzzy} of the fuzzy-c-means \citep{dias2019fuzzy} python package. Clustering performance is evaluated using the Normalized Mutual Information (NMI) score \citep{yang2016comparative}.

Four widely used data sets from the field of bioinformatics are employed along with two popular data sets benchmark data set for text and image clustering, respectively:

\noindent
\begin{minipage}{.44\textwidth}
    \begin{itemize}
        \setlength{\itemsep}{5pt}
        \setlength{\parskip}{0pt}
        \setlength{\parsep}{0pt}
        \item the Deng, \citep{DengData}
        \item the TGCA Pan-cancer\footnotemark (Cancer),  
        \item the USPS, \citep{291440}
    \end{itemize}    
\end{minipage}
\hfill
\begin{minipage}{.55\textwidth}
    \begin{itemize}
        \setlength{\itemsep}{5pt}
        \setlength{\parskip}{0pt}
        \setlength{\parsep}{0pt}
        \item the Baron, \citep{baron2016single}
        \item the Chen, \citep{chen2017single}
        \item the BBC, \citep{greene06icml},
        \vspace{-3mm}
    \end{itemize}    
\end{minipage}{}
\footnotetext{\url{https://www.doi.org/10.7303/syn300013}}

All experiments took place on a server computer with Linux operating system, kernel version 5.11.0, with an Intel Core i7-10700K CPU @ 3.80GHz and four DDR4 RAM dims of 32GB with 2133MHz frequency. Default parameters were used for the execution of all the algorithms, and the actual number of clusters was provided to algorithms as a parameter when required.

In Table \ref{tbl:resutls} we present the mean performance of all methods with respect to execution time (time in secs) and NMI across 100 experiments. We observe that HiPart implementations perform exquisitely in terms of execution time while still being comparable with respect to clustering performance.

\noindent
\begin{minipage}[b]{0.52\textwidth}
    \centering
    \scriptsize
    \begin{tabular}{|l|cc|cc|c|}
\hline
Algorithm & time    & NMI               & time      & NMI             & \\
\hline\hline
          & \multicolumn{2}{c|}{Deng (135, 12548)}   & \multicolumn{2}{c|}{Baron (1886, 14878)} & \parbox[t]{2mm}{\multirow{21}{*}{\rotatebox[origin=c]{90}{Gene Expression Data}}} \\
\cline{1-5}
iPDDP     & 0.10    & 0.76              & 1.16      & 0.08            & \\
dePDDP    & 0.14    & 0.70              & 2.16      & 0.53            & \\
PDDP      & 0.15    & 0.54              & 2.55      & 0.53            & \\
kM-PDDP   & 0.25    & 0.61              & 3.81      & 0.52            & \\
BKM       & 0.50    & 0.64              & 11.51     & 0.52            & \\
k-Means   & 0.14    & 0.71              & 7.52      & 0.48            & \\
AGG       & 0.04    & 0.72              & 14.54     & 0.51            & \\
OPTICS    & 27.78   & 0.48              & 710.99    & 0.13            & \\
FCM       & 1.37    & 0.68              & 163.63    & 0.45            & \\
\cline{1-5}
          & \multicolumn{2}{c|}{Cancer (801, 20531)} & \multicolumn{2}{c|}{Chen (14437, 23284)}& \\
\cline{1-5}
iPDDP     & 0.90    & 0.67              & 13.18     & 0.30            & \\
dePDDP    & 1.06    & 0.93              & 20.71     & 0.36            & \\
PDDP      & 1.04    & 0.74              & 37.52     & 0.48            & \\
kM-PDDP   & 1.27    & 0.86              & 53.73     & 0.48            & \\
BKM       & 5.94    & 0.88              & 255.85    & 0.48            & \\
k-Means   & 1.54    & 0.98              & 249.72    & 0.48            & \\
AGG       & 3.09    & 0.98              & 1218.68   & 0.49            & \\
OPTICS    & 266.39  & 0.34              & 27089.35  & 0.00            & \\
FCM       & 5.53    & 0.53              & 5710.83   & 0.26            & \\
\hline\hline
          & \multicolumn{2}{c|}{USPS (4575, 256)}    & \multicolumn{2}{c|}{BBC (2225, 21213)} & \parbox[t]{2mm}{\multirow{10}{*}{\rotatebox[origin=c]{90}{Benchmark Data}}} \\
\cline{1-5}
iPDDP     & 0.02    & 0.55              & 2.02      & 0.60            & \\
dePDDP    & 0.05    & 0.65              & 1.70      & 0.60            & \\
PDDP      & 0.04    & 0.60              & 2.57      & 0.78            & \\
kM-PDDP   & 0.17    & 0.50              & 2.93      & 0.74            & \\
BKM       & 0.38    & 0.58              & 9.92      & 0.65            & \\
k-Means   & 0.12    & 0.72              & 4.68      & 0.35            & \\
AGG       & 1.15    & 0.77              & 23.18     & 0.64            & \\
OPTICS    & 635.05  & 0.04              & 1055.75   & 0.06            & \\
FCM       & 2.92    & 0.58              & 4.73      & 0.60            & \\
\hline
    \end{tabular}
    \captionof{table}{Clustering results with respect to execution time and clustering performance.}
    \label{tbl:resutls}
\end{minipage}
\hfill
\begin{minipage}[b]{0.46\textwidth}
    \centering
    \scriptsize
    \includegraphics[width=\textwidth, height=3.18in]{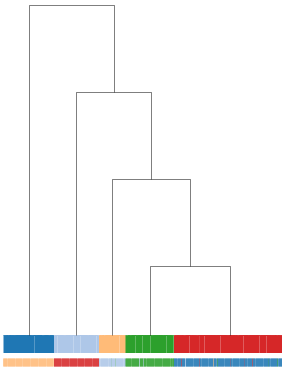}
    \captionof{figure}{Dendrogram figure for the Cancer data set with the use of the dePDDP algorithm and the dendrogram visualization module of the HiPart library. The line below the tree represents the colour of the original cluster each sample belongs.}
    \label{fig:dendrogram}
\end{minipage}
\section{Conclusions and Future Work}
We present a highly time-efficient clustering package with a suite of tools that give the capability of addressing problems in high-dimensional data clustering. Also, the developed new visualization tools enhance understanding and identification of the underlying clustering data structure.

We plan to continuously expand the HiPart package in the future through the addition of more hierarchical algorithms and by providing even more options for dimensionality reduction, such as the use of recent projection pursuit methodologies \citep{pavlidis2016minimum, hofmeyr2016clustering, hofmeyr2019minimum, hofmeyr2019ppci}. Our final aim is to establish the golden standard when considering hierarchical divisive clustering.

\acks{This project has received funding from the Hellenic Foundation for Research and Innovation (HFRI), under grant agreement No 1901.}

\bibliography{refs}

\end{document}